# Towards a unified theory of logic programming semantics: Level mapping characterizations of selector generated models

Pascal Hitzler[1*] and Sibylle Schwarz[2]

[1] AIFB, Universität Karlsruhe (TH)
email: hitzler@aifb.uni-karlsruhe.de
[2] Institut für Informatik, Martin-Luther-Universität Halle-Wittenberg
email: schwarzs@informatik.uni-halle.de

**Abstract.** Currently, the variety of expressive extensions and different semantics created for logic programs with negation is diverse and heterogeneous, and there is a lack of comprehensive comparative studies which map out the multitude of perspectives in a uniform way. Most recently, however, new methodologies have been proposed which allow one to derive uniform characterizations of different declarative semantics for logic programs with negation. In this paper, we study the relationship between two of these approaches, namely the level mapping characterizations due to [17], and the selector generated models due to [24]. We will show that the latter can be captured by means of the former, thereby supporting the claim that level mappings provide a very flexible framework which is applicable to very diversely defined semantics.

## 1 Introduction

Applications of logic programming in intelligent systems, knowledge management, semantic web, and elsewhere, necessitate the extension of the core Horn paradigm by expressive features such as non-monotonic negation, disjunctive consequences, fuzziness, dynamic updating, aggregates, etc. As logic programming is a declarative paradigm, it is of vital importance to provide model-theoretic semantics for such extensions. This is usually done, but rarely in a systematic way. Often, an existing semantics for a related syntactic paradigm is transferred to an analogous semantics on the new paradigm, guided not by systematic studies or results, but rather by the problem domain considered and by intuitive insights into the knowledge modelling aspects.

As a result of this, there is a plethora of different proposals for semantics which are somehow related, but whose exact relationships are rarely studied. The lack of reconciliating work makes it difficult for students and young researcher to

* The first named author acknowledges support by the German Federal Ministry of Education and Research under the SmartWeb project, and by the Deutsche Forschungsgemeinschaft (DFG) under the ReaSem project.

obtain a coherent picture of the subject area, and is a hindrance for a systematic advance within the research community.

In order to address the need for charting the existing semantic landscape, a methodology has been proposed in [17] for characterizing different semantics in a uniform way. The approach is very flexible and allows to cast semantics of very different origin and style into uniform characterizations using level mappings, i.e. mappings from atoms to ordinals, in the spirit of the definition of acceptable programs [2], the use of stratification [1,22] and a characterization of stable models by Fages [8]. These characterizations display syntactic and semantic dependencies between language elements by means of the preorders on ground atoms induced by the level mappings, and thus allow inspection of and comparison between different semantics, as exhibited in [14,17].

For the syntactically restricted class of normal logic programs, the most important semantics — and some others — have already been characterized and compared, and this was spelled out in [14,17]. Due to the inherent flexibility of the framework, it is clear that studies of extended syntax are also possible, but have so far not been carried out. In this paper, we will present a non-trivial technical result which provides a first step towards a comprehensive comparative study of different semantics for logic programs under extended syntax.

More precisely, among the many proposals for semantics for logic programs under extended syntax we will study a very general approach due to Schwarz [23,24]. In this framework, arbitrary formulae are allowed in rule heads and bodies, and it encompasses the inflationary semantics [18], the stable semantics for normal and disjunctive programs [11,21], and the stable generated semantics [13]. It can itself be understood as a unifying framework for different semantics.

In this paper, we will provide a single theorem – and some corollaries thereof – which gives a characterization of general selector generated models by means of level mappings. It thus provides a link between these two frameworks, and implicitly yields level mapping characterizations of the semantics encompassed by the selector generated approach.

The plan of the paper is as follows. In Section 2 we will fix preliminaries and notation. In Section 3 we will review selector generated models as introduced in [23,24], and in Section 4 we recall main notions and results about level mapping characterizations as studied in [14,17]. In Section 5, we present our main result, Theorem 5, which gives a level-mapping characterization of general selector generated models. In Section 6 we study corollaries from Theorem 5 concerning specific cases of interest encompassed by the result. Section 7 presents some related work. We eventually conclude and discuss further work in Section 8.

This paper is a substantially revised and extended version of [16].

## 2 Preliminaries

Throughout the paper, we will consider a language $\mathcal{L}$ of propositional logic over some set of propositional variables, or *atoms*, $A$, and connectives $\Sigma^{cl} =$ $\{\neg, \vee, \wedge, \mathbf{t}, \mathbf{f}\}$, as usual. A *rule* $r$ is a pair of formulae from $\mathcal{L}$ denoted by $\varphi \Rightarrow \psi$. $\varphi$ is called the *body* of the rule, denoted by $\mathsf{body}(r)$, and $\psi$ is called the *head* of the rule, denoted by $\mathsf{head}(r)$. A *program* is a set of rules.[1] A *literal* is an atom or a negated atom, and $\mathsf{Lit}\,(A)$ denotes the set of all literals in $\mathcal{L}$. For a set of connectives $C \subseteq \Sigma^{cl}$ we denote by $\mathsf{Lg}\,(C, A)$ the set of all formulae over $\mathcal{L}$ in which only connectives from $C$ occur.

**Table 1.** Notions of specific types of rules.

| *rule is called* | *set* | *condition* |
|---|---|---|
| *definite* | LP | $\mathsf{body}(r) \in \mathsf{Lg}\left(\{\wedge, \mathbf{t}\}, A\right)$ and $\mathsf{head}(r) \in A$ |
| *normal* | NLP | $\mathsf{body}(r) \in \mathsf{Lg}\left(\{\wedge, \mathbf{t}\}, \mathsf{Lit}\,(A)\right)$ and $\mathsf{head}(r) \in A$ |
| *head-atomic* | HALP | $\mathsf{body}(r) \in \mathsf{Lg}\left(\Sigma^{cl}, A\right)$ and $\mathsf{head}(r) \in A$ |
| *pos. head disj.* | $\mathsf{DLP}^+$ | $\mathsf{body}(r) \in \mathsf{Lg}\left(\{\wedge, \mathbf{t}\}, \mathsf{Lit}\,(A)\right)$ and $\mathsf{head}(r) \in \mathsf{Lg}\left(\{\vee\}, A\right)$ |
| *disjunctive* | DLP | $\mathsf{body}(r) \in \mathsf{Lg}\left(\{\wedge, \mathbf{t}\}, \mathsf{Lit}\,(A)\right)$, $\mathsf{head}(r) \in \mathsf{Lg}\left(\{\vee, \mathbf{f}\}, \mathsf{Lit}\,(A)\right)$ |
| *head-disjunctive* | HDLP | $\mathsf{body}(r) \in \mathsf{Lg}\left(\Sigma^{cl}, A\right)$, $\mathsf{head}(r) \in \mathsf{Lg}\left(\{\vee, \mathbf{f}\}, \mathsf{Lit}\,(A)\right)$ |
| *generalized* | GLP | no condition |

Further terminology is introduced in Table 1. The abbreviations in the second column denote the sets of all rules with the corresponding property. A program containing only definite (normal, etc.) rules is called *definite* (*normal*, etc.). Programs not containing the negation symbol $\neg$ are called *positive*. *Facts* are rules $r$ where $\mathsf{body}(r) = \mathbf{t}$, denoted by $\Rightarrow \mathsf{head}(r)$.

The *base* $\mathsf{B}_P$ is the set of all atoms occurring in a program $P$. A two-valued *interpretation* of a program $P$ is represented by a subset of $\mathsf{B}_P$, as usual. By $\mathbf{I}_P$ we denote the set of all interpretations of $P$. It is a complete lattice with respect to the subset ordering $\subseteq$. For an interpretation $I \in \mathbf{I}_P$, we define $\uparrow I = \{J \in \mathbf{I}_P \mid I \subseteq J\}$ and $\downarrow I = \{J \in \mathbf{I}_P \mid J \subseteq I\}$. Sets of the form $[I, J] = \uparrow I \cap \downarrow J$ are called *intervals* of interpretations.

The model relation $M \models \varphi$ for an interpretation $M$ and a propositional formula $\varphi$ is defined as usual in propositional logic, and $\mathrm{Mod}(\varphi)$ denotes the set of all models of $\varphi$. Two formulae $\varphi$ and $\psi$ are *logically equivalent*, written $\varphi \equiv \psi$, iff $\mathrm{Mod}(\varphi) = \mathrm{Mod}(\psi)$.

A formula $\varphi$ is *satisfied* by a set $\mathbf{J} \subseteq \mathbf{I}_P$ of interpretations if each interpretation $J \in \mathbf{J}$ is a model of $\varphi$. For a program $P$, a set $\mathbf{J} \subseteq \mathbf{I}_P$ of interpretations determines the set of all rules which *fire* under $\mathbf{J}$, formally $\mathsf{fire}(P, \mathbf{J}) = \{r \in P \mid \forall J \in \mathbf{J} : J \models \mathsf{body}(r)\}$. An interpretation $M$ is called a *model* of a rule $r$ (or *satisfies* $r$) if $M$ is a model of the formula $\neg\mathsf{body}(r) \vee \mathsf{head}(r)$. An interpretation $M$ is a *model* of a program $P$ if it satisfies each rule in $P$.

For conjunctions or disjunctions $\varphi$ of literals, $\varphi^+$ denotes the set of all atoms occurring positively in $\varphi$, and $\varphi^-$ contains all atoms that occur negated in $\varphi$. For instance, for the formula $\varphi = (a \wedge \neg b \wedge \neg a)$ we have $\varphi^+ = \{a\}$ and $\varphi^- = \{a, b\}$. In heads $\varphi$ consisting only of disjunctions of literals, we always assume without loss of generality that $\varphi^+ \cap \varphi^- = \emptyset$.

[1] Note that infinite sets of rules are allowed. This is important because it means that the notion of *program* encompasses first-order logic programs with function symbols, as they can be transformed into (infinite) propositional programs by grounding, as usual.

If $\varphi$ is a *conjunction* of literals, we abbreviate $M \models \bigwedge_{a \in \varphi^+} a$ (i.e. $\varphi^+ \subseteq M$) by $M \models \varphi^+$, and $M \models \bigwedge_{a \in \varphi^-} \neg a$ (i.e. $\varphi^- \cap M = \emptyset$) by $M \models \varphi^-$, abusing notation. If $\varphi$ is a *disjunction* of literals, we write $M \models \varphi^+$ for $M \models \bigvee_{a \in \varphi^+} a$ (i.e. $M \cap \varphi^+ \neq \emptyset$), and $M \models \varphi^-$ for $M \models \bigvee_{a \in \varphi^-} \neg a$ (i.e. $\varphi^- \not\subseteq M$).

By iterative application of rules from a program $P \subseteq \mathsf{GLP}$ starting in the least interpretation $\emptyset \in \mathbf{I}_P$, we can create monotonically increasing (transfinite) sequences of interpretations of the program $P$, as follows.

**Definition 1.** *A (transfinite) sequence $C$ of length $\alpha$ of interpretations of a program $P \subseteq \mathsf{GLP}$ is called a $P$-*chain *iff*

(**C0**) $C_0 = \emptyset$,
(**C**$\beta$) $C_{\beta+1} \in \mathrm{Min}(\uparrow C_\beta \cap \mathrm{Mod}(\mathsf{head}(Q_\beta)))$ *for all $\beta$ with $\beta + 1 < \alpha$ and some set of rules $Q_\beta \subseteq P$, and*
(**C**$\lambda$) $C_\lambda = \bigcup\{C_\beta \mid \beta < \lambda\}$ *for all limit ordinals $\lambda < \alpha$.*

$\mathbf{C}_P$ *denotes the collection of all $P$-chains.*

Note that all $P$-chains increase monotonically with respect to $\subseteq$.

In the proof of Theorem 5, we will make use of the following straightforward lemma from [24].

**Lemma 1.** *For any set of interpretations $\mathbf{J} \subseteq \mathbf{I}_P$ and any interpretation $K \in \mathbf{I}_P$ we have* $\mathrm{Min}(\mathbf{J} \cap \downarrow K) = \mathrm{Min}(\mathbf{J}) \cap \downarrow K$. □

## 3 Selector generated models

In [23,24], a framework for defining declarative semantics of generalized logic programs was introduced, which encompasses several other semantics, as already mentioned in the introduction. Parametrization within this framework is done via so-called *selector functions*, defined as follows.

**Definition 2.** *A* selector *is a function* $\mathrm{Sel} : \mathbf{C}_P \times \mathbf{I}_P \to 2^{\mathbf{I}_P}$*, satisfying* $\emptyset \neq \mathrm{Sel}(C, I) \subseteq [I, \sup(C)]$ *for all $P$-chains $C$ and each interpretation $I \in \downarrow \sup(C)$.*

*Example 1.* In this paper, we will have a closer look at the following selectors.

| | |
|---|---|
| lower bound selector | $\mathrm{Sel}_{\mathsf{l}}(C, I) = \{I\}$ |
| lower and upper bound selector | $\mathrm{Sel}_{\mathsf{lu}}(C, I) = \{I, \sup(C)\}$ |
| interval selector | $\mathrm{Sel}_{\mathsf{i}}(C, I) = [I, \sup(C)]$ |
| chain selector | $\mathrm{Sel}_{\mathsf{c}}(C, I) = [I, \sup(C)] \cap C$ |

We use selectors Sel to define nondeterministic successor functions $\Omega_P$ on $\mathbf{I}_P$, as follows.

**Definition 3.** *Given a selector* Sel : $\mathbf{C}_P \times \mathbf{I}_P \to 2^{\mathbf{I}_P}$ *and a program* $P$*, the function* $\Omega_P : (\mathbf{C}_P \times \mathbf{I}_P \to 2^{\mathbf{I}_P}) \times \mathbf{C}_P \times \mathbf{I}_P \to 2^{\mathbf{I}_P}$ *is defined by*

$$\Omega_P(\mathrm{Sel}, C, I) = \mathrm{Min}\left([I, \sup(C)] \cap \mathrm{Mod}\left(\mathsf{head}\left(\mathsf{fire}\left(P, \mathrm{Sel}\left(C, I\right)\right)\right)\right)\right).$$

$\Omega_P(\mathrm{Sel}, C, \cdot)$ simulates a step in a reasonig process guided by $P$ with the goal to achieve the interpretation $\sup(C)$ along a chain $C$ of interpretations. All rules whose bodies are satisfied in all interpretations chosen by $\mathrm{Sel}(C, I)$ are applied simultaneously to construct potential successor interpretations of $I$. To keep $\Omega_P(\mathrm{Sel}, C, \cdot)$ monotonic in the last argument, we discard all interpretations $J$ where $J \subset I$ or $\sup(C) \subset J$ and collect the minimal among all remaining interpretations in $\Omega_P(\mathrm{Sel}, C, I)$.

With the first two arguments (the selector Sel and the chain $C$) fixed, the function $\Omega_P(\mathrm{Sel}, C, I)$ can be understood as a nondeterministic consequence operator. Iteration of the function $\Omega_P(\mathrm{Sel}, C, \cdot)$ from the least interpretation $\emptyset$ creates sequences of interpretations. This leads to the following definition of $(P, M, \mathrm{Sel})$-chains.

**Definition 4.** *A* $(P, M, \mathrm{Sel})$-chain *is a* $P$*-chain satisfying*

($\mathbf{C}\sup$) $M = \sup(C)$ *and*
($\mathbf{C}\beta_{\mathrm{Sel}}$) $C_{\beta+1} \in \Omega_P(\mathrm{Sel}, C, C_\beta)$ *for all* $\beta$*, where* $\beta + 1 < \kappa$ *and* $\kappa$ *is the length of the transfinite sequence* $C$*.*

Thus, $(P, M, \mathrm{Sel})$-chains are monotonically increasing sequences $C$ of interpretations of $P$, that reproduce themselves by iterating $\Omega_P$. Note that this definition is non-constructive.

The main concept of the selector semantics is fixed in the following definition.

**Definition 5.** *A model* $M$ *of a program* $P \subseteq \mathsf{GLP}$ *is* Sel-generated *if and only if there exists a* $(P, M, \mathrm{Sel})$*-chain* $C$*. The* Sel*-semantics of the program* $P$ *is the set* $\mathrm{Mod}_{\mathrm{Sel}}(P)$ *of all* Sel*-generated models of* $P$*.*

*Example 2.* The program $P$ consisting of the rules

$$\Rightarrow a \tag{1}$$
$$a \Rightarrow b \tag{2}$$
$$(a \vee \neg c) \wedge (c \vee \neg a) \Rightarrow c \tag{3}$$

has the only $\mathrm{Sel}_{\mathsf{l}}$-generated model $\{a, b, c\}$, namely via the chain $C_1 = (\emptyset \overset{1,3}{\to} \{a, c\} \overset{2}{\to} \{a, b, c\})$), where the rules applied in each step are denoted above the arrows. $\{a, b\}$ and $\{a, b, c\}$ are $\mathrm{Sel}_{\mathsf{lu}}$-generated (and $\mathrm{Sel}_{\mathsf{c}}$-generated) models, namlely via the chains $C_2 = (\emptyset \overset{1}{\to} \{a\} \overset{2}{\to} \{a, b\})$ and $C_1$). $\{a, b\}$ is the only $\mathrm{Sel}_{\mathsf{i}}$-generated model of $P$, namely via $C_2$.

Some properties of semantics generated by the selectors in Example 1 were studied in [23,24]. For all $* \in \{\mathsf{l}, \mathsf{lu}, \mathsf{i}, \mathsf{c}\}$, we abbreviate $\mathrm{Mod}_{\mathrm{Sel}_*}$ by $\mathrm{Mod}_*(P)$ (for instance $Mod_{\mathrm{Sel}_{\mathsf{lu}}}$ by $\mathrm{Mod}_{\mathsf{lu}}(P)$). In Section 6, we will make use of the following results.

**Theorem 1** **([24]).**

1. *For definite programs* $P \subseteq \mathsf{LP}$*, the unique element contained in* $\mathrm{Mod}_{\mathsf{l}}(P) = \mathrm{Mod}_{\mathsf{lu}}(P) = \mathrm{Mod}_{\mathsf{c}}(P) = \mathrm{Mod}_{\mathsf{i}}(P)$ *is the* least *model of* $P$*.*
2. *For normal programs* $P \subseteq \mathsf{NLP}$*, the unique element of* $\mathrm{Mod}_{\mathsf{l}}(P)$ *is the* inflationary *model of* $P$ *(as introduced in [18]).*
3. *For normal programs* $P \subseteq \mathsf{NLP}$*, the set* $\mathrm{Mod}_{\mathsf{lu}}(P) = \mathrm{Mod}_{\mathsf{c}}(P) = \mathrm{Mod}_{\mathsf{i}}(P)$ *contains exactly all* stable *models of* $P$ *(as defined in [11]).*
4. *For positive head disjunctive programs* $P \subseteq \mathsf{DLP}^{+}$*, the minimal elements in* $\mathrm{Mod}_{\mathsf{lu}}(P) = \mathrm{Mod}_{\mathsf{c}}(P) = \mathrm{Mod}_{\mathsf{i}}(P)$ *are exactly all stable models of* $P$ *(as defined in [21]), but for generalized programs* $P \subseteq \mathsf{GLP}$*, the sets* $\mathrm{Mod}_{\mathsf{lu}}(P)$*,* $\mathrm{Mod}_{\mathsf{c}}(P)$*, and* $\mathrm{Mod}_{\mathsf{i}}(P)$ *may differ.*
5. *For generalized programs* $P \subseteq \mathsf{GLP}$*,* $\mathrm{Mod}_{\mathsf{i}}(P)$ *is the set of* stable generated models *of* $P$ *(as defined in [13]).* □

Theorem 1 shows that the framework of selector semantics covers some of the most important declarative semantics for normal logic programs. Further characterizations of semantics can be found in [24], where the approach has also been lifted to many-valued logics and corresponding semantics, e.g. well-founded and paraconsistent semantics. Selector generated models thus provide a consistent unifying framework which also allows to derive natural extension of known or new semantics to generalized logic programs and enables systematic comparisons of new and known semantics.

## 4 Level mapping characterizations

In [14,17], a uniform approach to different semantics for logic programs was given which has been developed independently to that of selector generated models. It is based on the notion of *level mapping*, as follows.

**Definition 6.** *A* level mapping *for a logic program* $P \subseteq \mathsf{GLP}$ *is a function* $l : \mathsf{B}_P \to \alpha$*, where* $\alpha$ *is an ordinal.*[2]

The general goals of level characterizations of semantics are similar to those for studying selector generated models: to obtain a unifying framework which encompasses different semantics, in order to make them comparable, to reconcile the many heterogeneous approaches to logic programming semantics, and to provide general guidance for further developments in the field. In order to display the style of level-mapping characterizations for semantics, we recall some examples which we will further discuss later on.

**Theorem 2** **([17]).** *Every definite program* $P \subseteq \mathsf{LP}$ *has exactly one model* $M$*, such that there exists a level mapping* $l : \mathsf{B}_P \to \alpha$ *satisfying*

[2] Note that transfinite ordinals are needed in order to treat infinite programs. Cf. also Footnote 1.

**(Fd)** *for every atom* $a \in M$ *there exists a rule* $\bigwedge_{b \in B} b \Rightarrow a \in P$ *such that* $B \subseteq M$ *and* $\max\{l(b) \mid b \in B\} < l(a)$.

*Furthermore,* $M$ *coincides with the least model of* $P$. □

*Example 3.* Let $P$ be the ground instantiation of the program consisting of the rules

$$\Rightarrow p(0)$$
$$p(X) \Rightarrow p(s(X))$$

where $X$ denotes a variable and 0 a constant symbol. Write $s^n(0)$ for the term $s(\dots s(0)\dots)$ in which the symbol $s$ appears $n$ times. Then $\{p(s^n(0)) \mid n \in \mathbb{N}\}$ is the least model of $P$. A level mapping corresponding to Theorem 2 is given by $l(p(s^n(0))) = n$ for all $n \in \mathbb{N}$.

The following theorem is actually due to Fages.

**Theorem 3 ([8]).** *Let* $P$ *be a normal program and* $M$ *be a model for* $P$*. Then* $M$ *is a stable model of* $P$ *iff there exists a level mapping* $l : \mathsf{B}_P \to \alpha$ *satisfying*

**(Fs)** *for each atom* $a \in M$ *there exists a rule* $r \in P$ *with* $\mathsf{head}(r) = a$, $\mathsf{body}(r)^+ \subseteq M$, $\mathsf{body}(r)^- \cap M = \emptyset$, *and* $\max\{l(b) \mid b \in \mathsf{body}(r)^+\} < l(a)$. □

*Example 4.* Let $P$ be the program consisting of the following rules.

$$q \Rightarrow s$$
$$\neg p \Rightarrow q$$
$$p \Rightarrow p$$

For the stable model $\{s, q\}$ of $P$ a level mapping $l$ corresponding to Theorem 3 satisfies $l(q) = 0$ and $l(s) = 1$, while $l(p)$ can be an arbitrary value.

It is evident, that among the level mappings satisfying the respective conditions in Theorems 2 and 3, there exist pointwise minimal ones.

The paper [17], where level mapping characterizations were originally introduced, covers further semantics including the well-founded [25] and the Fitting semantics [9]. It also provides a general proof scheme for obtaining level mapping characterizations.

The following result and example are taken from [15].

**Theorem 4.** *Let* $P$ *be a positive head disjunctive program. Then a model* $M$ *of* $P$ *is a disjunctive stable model of* $P$ *if and only if there exists a total level mapping* $l : \mathsf{B}_P \to \alpha$ *such that for each* $a \in M$ *there exists a rule* $r$ *in* $P$ *with* $\mathsf{body}(r)^+ \subseteq M$, $\mathsf{body}(r)^- \cap M = \emptyset$, $\mathsf{head}(r)^+ \cap M = \{a\}$ *and* $\max\{l(b) \mid b \in \mathsf{body}(r)^+\} < l(a)$. □

*Example 5.* Let $P$ be the following disjunctive program:

$$\begin{aligned}\neg c &\Rightarrow a \vee b\\ \neg a \wedge \neg b &\Rightarrow c\\ a &\Rightarrow d \vee e\\ d \wedge \neg e &\Rightarrow f\end{aligned}$$

Program $P$ has the four (disjunctive) stable models $\{b\}, \{c\}, \{a, d, f\}$ and $\{a, e\}$. Corresponding level mappings as in Theorem 4 are $l(a) = l(b) = l(c) = l(d) = l(e) = l(f) = 0$ for $\{b\}$ and $\{c\}$, $l(a) = l(b) = l(c) = l(e) = 0$, $l(d) = 1$,$l(f) = 2$ for $\{a, d, f\}$ and $l(a) = l(b) = l(c) = l(d) = l(f) = 0$, $l(e) = 1$ for $\{a, e\}$.

## 5 Selector generated models via level mappings

We set out to prove a general theorem which characterizes selector generated models by means of level mappings, in the style of the results displayed in Section 4. The following notion will ease notation considerably.

**Definition 7.** *For a level mapping $l : \mathsf{B}_P \to \alpha$ for a program $P \subseteq \mathsf{GLP}$ and an interpretation $M \subseteq \mathsf{B}_P$, the elements of the (transfinite) sequence $\mathsf{C}^{l,M}$ consisting of interpretations of $P$ are for all $\beta < \alpha$ defined by*

$$\mathsf{C}^{l,M}_\beta = \{a \in M \mid l(a) < \beta\} = M \cap \bigcup_{\gamma<\beta} l^{-1}(\gamma).$$

*Remark 1.* Definition 7 implies that

1. the (transfinite) sequence $\mathsf{C}^{l,M}$ is monotonically increasing,
2. $\mathsf{C}^{l,M}_0 = \emptyset$, and
3. $M = \bigcup_{\beta<\alpha} \mathsf{C}^{l,M}_\beta = \sup \mathsf{C}^{l,M}$.

The following theorem provides a translation between the definition of selector semantics and a level mapping characterization. This theorem and its generalization in Corollary 1 are the main results of this paper.

**Theorem 5.** *Let $P \subseteq \mathsf{HDLP}$ be a head disjunctive program and $M \in \mathbf{I}_P$. Then $M$ is a* Sel*-generated model of $P$ iff there exists a level mapping $l : \mathsf{B}_P \to \alpha$ satisfying the following properties.*

**(L1)** $M = \sup\left(\mathsf{C}^{l,M}\right) \in \mathrm{Mod}\,(P)$.
**(L2)** *For all $\beta$ with $\beta + 1 < \alpha$ we have*

$$\mathsf{C}^{l,M}_{\beta+1} \setminus \mathsf{C}^{l,M}_\beta \in \mathrm{Min}\left\{ J \in \mathbf{I}_P \,\middle|\, J \models \mathsf{head}\left(R\left(\mathsf{C}^{l,M}_\beta, J\right)\right)^+ \right\}, \quad \textit{where}$$

$$R\left(\mathsf{C}^{l,M}_\beta, J\right) = \left\{ r \in \mathsf{fire}\left(P, \mathrm{Sel}\left(\mathsf{C}^{l,M}, \mathsf{C}^{l,M}_\beta\right)\right) \,\middle|\, \begin{array}{l} \mathsf{C}^{l,M}_\beta \not\models \mathsf{head}\,(r)^+ \;\textit{ and} \\ J \cup \mathsf{C}^{l,M}_\beta \not\models \mathsf{head}\,(r)^- \end{array} \right\}.$$

**(L3)** *For all limit ordinals* $\lambda < \alpha$ *we have* $\mathsf{C}_{\lambda}^{l,M} = \bigcup_{\beta<\lambda} \mathsf{C}_{\beta}^{l,M}$.

*Example 6.* For the program in Example 2 level mappings corresponding to Theorem 5 satisfy $l_1(a) = l_1(b) = 0$ and $l_1(b) =$ for $\mathrm{Sel}_{\mathsf{l}}$, and $l_2(a) = 0$, $l_2(b) = l_2(c) = 1$ for $\mathrm{Sel}_{\mathsf{i}}$. For the two $\mathrm{Sel}_{\mathsf{lu}}$-generated models the level mappings $l_1$ respectively $l_2$ can also be used.

*Remark 2.* As $P$ is a head disjunctive program, we have $\mathsf{C}_{\beta}^{l,M} \not\models \mathsf{head}(r)^+$ iff $\mathsf{head}(r)^+ \cap \mathsf{C}_{\beta}^{l,M} = \emptyset$, and $J \cup \mathsf{C}_{\beta}^{l,M} \not\models \mathsf{head}(r)^-$ iff $\mathsf{head}(r)^- \subseteq J \cup \mathsf{C}_{\beta}^{l,M}$, thus

$$R\left(\mathsf{C}_{\beta}^{l,M}, J\right) = \left\{ r \in \mathsf{fire}\left(P, \mathrm{Sel}\left(\mathsf{C}^{l,M}, \mathsf{C}_{\beta}^{l,M}\right)\right) \,\middle|\, \begin{array}{l} \mathsf{head}(r)^+ \cap \mathsf{C}_{\beta}^{l,M} = \emptyset \text{ and} \\ \mathsf{head}(r)^- \subseteq J \cup \mathsf{C}_{\beta}^{l,M} \end{array} \right\}.$$

Also note that for every rule $r \in \mathsf{fire}\left(P, \mathrm{Sel}\left(\mathsf{C}^{l,M}, \mathsf{C}_{\beta}^{l,M}\right)\right) \setminus R\left(\mathsf{C}_{\beta}^{l,M}, J\right)$, we have $\downarrow\left(\mathsf{C}_{\beta}^{l,M} \cup J\right) \subseteq \mathrm{Mod}\left(\mathsf{head}(r)^-\right)$ or $\uparrow \mathsf{C}_{\beta}^{l,M} \subseteq \mathrm{Mod}\left(\mathsf{head}(r)^+\right)$. Thus all of these rules are satisfied in the interval $\left[\mathsf{C}_{\beta}^{l,M}, \mathsf{C}_{\beta}^{l,M} \cup J\right]$.

*Proof.* (of Theorem 5)

First assume that $M$ is a Sel-generated model of $P$, and recall that by Definition 5, an interpretation $M$ is a Sel-generated model of $P$ iff there exists a $(P, M, \mathrm{Sel})$-chain $C$. Let $\alpha$ be the length of $C$. $l(a) = \min\{\beta \mid a \in C_\beta\} - 1$ for all $a \in \mathsf{B}_P$. We show that this function $l$ satisfies **(L1)**,**(L2)** and **(L3)**.

We first show $\mathsf{C}^{l,M} = C$ for the sequence $\mathsf{C}^{l,M}$ determined by $l$ and $M$ according to Definition 7. From Remark 1, we know $\mathsf{C}_0^{l,M} = \emptyset$ and $\sup\left(\mathsf{C}^{l,M}\right) = M$. Moreover, for each $\beta < \alpha$, we have by definition of $l$ and Definition 7 $\mathsf{C}_{\beta}^{l,M} = \{a \in M \mid l(a) < \beta\} = \{a \in M \mid \min\{\gamma \mid a \in C_\gamma\} - 1 < \beta\} = C_\beta$. For all limit ordinals $\lambda < \alpha$ we have $\mathsf{C}_{\lambda}^{l,M} = \bigcup_{\beta<\lambda} \mathsf{C}_{\beta}^{l,M} = \bigcup_{\beta<\lambda} C_\beta = C_\lambda$. This proves $C = \mathsf{C}^{l,M}$. Since $C$ is a $(P, M, \mathrm{Sel})$-chain, it satisfies **(L1)** and **(L3)**.

It remains to show that $C$ satisfies **(L2)**. For all $\beta$ with $\beta + 1 < \alpha$, we show

(a) $C_{\beta+1} \setminus C_\beta \models \mathsf{head}\left(R\left(C_\beta, C_{\beta+1} \setminus C_\beta\right)\right)^+$ for

$$R\left(C_\beta, C_{\beta+1} \setminus C_\beta\right) = \left\{ r \in \mathsf{fire}\left(P, \mathrm{Sel}\left(C, C_\beta\right)\right) \,\middle|\, \begin{array}{l} C_\beta \not\models \mathsf{head}(r)^+ \text{ and} \\ C_\beta \cup C_{\beta+1} \setminus C_\beta \not\models \mathsf{head}(r)^- \end{array} \right\}$$
$$= \left\{ r \in \mathsf{fire}\left(P, \mathrm{Sel}\left(C, C_\beta\right)\right) \mid C_\beta \cap \mathsf{head}(r)^+ = \emptyset \text{ and } \mathsf{head}(r)^- \subseteq C_{\beta+1} \right\}$$

(b) $J \subseteq C_{\beta+1} \setminus C_\beta$ and $J \models \mathsf{head}\left(R\left(C_\beta, J\right)\right)^+$ implies $J = C_{\beta+1} \setminus C_\beta$.

We know $C_{\beta+1} \models \mathsf{head}\left(\mathsf{fire}\left(P, \mathrm{Sel}\left(C, C_\beta\right)\right)\right)$ since $C$ is a $(P, M, \mathrm{Sel})$-chain. For each $r \in R\left(C_\beta, C_{\beta+1} \setminus C_\beta\right)$, by $R\left(C_\beta, C_{\beta+1} \setminus C_\beta\right) \subseteq \mathsf{fire}\left(P, \mathrm{Sel}\left(C, C_\beta\right)\right)$ we have $C_{\beta+1} \models \mathsf{head}(r)$. By Remark 2, the set $R\left(C_\beta, C_{\beta+1} \setminus C_\beta\right)$ does not contain any rule $r \in \mathsf{fire}\left(P, \mathrm{Sel}\left(C, C_\beta\right)\right)$, where $C_{\beta+1} \models \mathsf{head}(r)$ is satisfied by $C_{\beta+1} \models \mathsf{head}(r)^-$ or $C_\beta \models \mathsf{head}(r)^+$, i.e. $\mathsf{head}(r)^+ \cap C_\beta \neq \emptyset$. Hence all rules $r$ from $R\left(C_\beta, C_{\beta+1} \setminus C_\beta\right) \subseteq \mathsf{fire}\left(P, \mathrm{Sel}\left(C, C_\beta\right)\right)$ satisfy $C_\beta \models \mathsf{head}(r)$ by $C_{\beta+1} \setminus C_\beta \cap \mathsf{head}(r)^+ \neq \emptyset$, i.e. $C_{\beta+1} \setminus C_\beta \models \mathsf{head}(r)^+$. This proves (a).

Now assume $J \subseteq C_{\beta+1} \setminus C_\beta$ and $J \models \mathsf{head}\,(R\,(C_\beta, J))^+$. We show $J \cup C_\beta \supseteq C_{\beta+1}$ which implies $J \supseteq C_{\beta+1} \setminus C_\beta$. First note that $J \cup C_\beta \subseteq [C_\beta, M] \cap \mathrm{Mod}\,(\mathsf{head}\,(\mathsf{fire}\,(P, \mathrm{Sel}\,(C, C_\beta))))$. Indeed $J \cup C_\beta \in {\uparrow} C_\beta$ is obvious and $J \cup C_\beta \in {\downarrow} M$ is implied by $J \subseteq C_{\beta+1} \setminus C_\beta$, i.e. $J \cup C_\beta \subseteq C_{\beta+1}$, and $C_{\beta+1} \subseteq M$ by monotonicity of the chain $C$. Now we show $J \cup C_\beta \models \mathsf{head}\,(\mathsf{fire}\,(P, \mathrm{Sel}\,(C, C_\beta)))$. Note first that all rules $r$ in the set $\mathsf{fire}\,(P, \mathrm{Sel}\,(C, C_\beta))$ satisfy one of the following conditions.

1. $C_\beta \cap \mathsf{head}\,(r)^+ \neq \emptyset$ and therefore $J \cup C_\beta \models \mathsf{head}\,(r)$ by $C_\beta \subseteq J \cup C_\beta$ or
2. $J \cup C_\beta \models \mathsf{head}\,(r)^-$ and therefore $J \cup C_\beta \models \mathsf{head}\,(r)$ or
3. none of 1. or 2. Then $r \in R\,(C_\beta, J)$ and due to $J \in \mathrm{Mod}\left(\mathsf{head}\,(R\,(C_\beta, J))^+\right)$ we have $J \cup C_\beta \models \mathsf{head}\,(r)^+$ and thus $J \cup C_\beta \models \mathsf{head}\,(r)$.

We can now conclude $J \cup C_\beta \supseteq C_{\beta+1}$ because $C_{\beta+1}$ is a minimal element of $[C_\beta, M] \cap \mathrm{Mod}\,(\mathsf{head}\,(\mathsf{fire}\,(P, \mathrm{Sel}\,(C, C_\beta))))$, which proves (b). Together, we have shown that $C_{\beta+1} \setminus C_\beta$ is a minimal element in $\left\{J \in \mathbf{I}_P \mid J \models \mathsf{head}\,(R\,(C_\beta, J))^+\right\}$, which shows that the level mapping $l$ satisfies **(L2)**. This finishes the first part of the proof.

For the converse, we show that for every level mapping $l$ for a program $P$ and an interpretation $M$ satisfying **(L1)**,**(L2)** and **(L3)** the sequence $\mathsf{C}^{l,M}$ is a $(P, M, \mathrm{Sel})$-chain. Let $l : \mathsf{B}_P \to \alpha$ be a level mapping and $M$ an interpretation for a program $P$. According to Definition 4, we show that the the sequence $\mathsf{C}^{l,M}$ satisfies:

($\mathbf{C0}$) $\mathsf{C}^{l,M}_0 = \emptyset$ (holds obviously by Remark 1.),
($\mathbf{C}\lambda$) $\mathsf{C}^{l,M}_\lambda = \bigcup\left\{\mathsf{C}^{l,M}_\beta \mid \beta < \lambda\right\}$ for all limit ordinals $\lambda < \alpha$,
($\mathbf{C}\,\sup$) $M = \bigcup\left\{\mathsf{C}^{l,M}_\beta \mid \beta < \alpha\right\} = \sup \mathsf{C}^{l,M}$ and
($\mathbf{C}\beta_{\mathrm{Sel}}$) $\mathsf{C}^{l,M}_{\beta+1} \in \mathit{\Omega}_P\left(\mathrm{Sel}, \mathsf{C}^{l,M}, \mathsf{C}^{l,M}_\beta\right)$ for all $\beta$ with $\beta + 1 < \alpha$.

By Remark 1 we know that $\mathsf{C}^{l,M}$ increases monotonically. By condition **(L1)** we have $M = \sup\left(\mathsf{C}^{l,M}\right) \in \mathrm{Mod}\,(P)$, i.e. ($\mathbf{C}\,\sup$), and condition **(L3)** implies ($\mathbf{C}\lambda$). It remains to show ($\mathbf{C}\beta_{\mathrm{Sel}}$), i.e. for all $\beta$ with $\beta + 1 < \alpha$ holds
$\mathsf{C}^{l,M}_{\beta+1} \in \mathrm{Min}\left(\left[\mathsf{C}^{l,M}_\beta, M\right] \cap \mathbf{M}\right)$ for $\mathbf{M} = \mathrm{Mod}\left(\mathsf{head}\left(\mathsf{fire}\left(P, \mathrm{Sel}\left(\mathsf{C}^{l,M}, \mathsf{C}^{l,M}_\beta\right)\right)\right)\right)$.
By Lemma 1 and monotonicity of $\mathsf{C}^{l,M}$ (i.e. $\mathsf{C}^{l,M}_{\beta+1} \in {\downarrow} M$), it suffices to show $\mathsf{C}^{l,M}_{\beta+1} \in \mathrm{Min}\left({\uparrow} \mathsf{C}^{l,M}_\beta \cap \mathbf{M}\right)$. We proceed in two steps:
(a) $\mathsf{C}^{l,M}_{\beta+1} \in {\uparrow} \mathsf{C}^{l,M}_\beta \cap \mathbf{M}$ (i.e. $\mathsf{C}^{l,M}_{\beta+1} \in \mathbf{M}$ by monotonicity of $\mathsf{C}^{l,M}$).
(b) $J \subseteq \mathsf{C}^{l,M}_{\beta+1}$ and $J \in {\uparrow} \mathsf{C}^{l,M}_\beta \cap \mathbf{M}$ implies $J = \mathsf{C}^{l,M}_{\beta+1}$.
Note that for every $r \in \mathsf{fire}\left(P, \mathrm{Sel}\left(\mathsf{C}^{l,M}, \mathsf{C}^{l,M}_\beta\right)\right)$ one of the following holds:
1. $\mathsf{C}^{l,M}_\beta \models \mathsf{head}\,(r)^+$, and therefore $\mathsf{C}^{l,M}_{\beta+1} \models \mathsf{head}\,(r)$ by $\mathsf{C}^{l,M}_\beta \subseteq \mathsf{C}^{l,M}_{\beta+1}$ or
2. $\mathsf{C}^{l,M}_{\beta+1} \models \mathsf{head}\,(r)^-$ and therefore $\mathsf{C}^{l,M}_{\beta+1} \models \mathsf{head}\,(r)$ or
3. none of 1. or 2. Then $r \in R\left(\mathsf{C}^{l,M}_\beta, \mathsf{C}^{l,M}_{\beta+1} \setminus \mathsf{C}^{l,M}_\beta\right)$ and thus $\mathsf{C}^{l,M}_{\beta+1} \models \mathsf{head}\,(r)$ by $\mathsf{C}^{l,M}_{\beta+1} \setminus \mathsf{C}^{l,M}_{\beta} \models \mathsf{head}\,(r)^{+}$ and condition **(L2)**.

Hence $\mathsf{C}^{l,M}_{\beta+1} \models \mathsf{head}\,(r)$ for each rule $r \in \mathsf{fire}\left(P, \mathrm{Sel}\left(\mathsf{C}^{l,M}, \mathsf{C}^{l,M}_{\beta}\right)\right)$ and thus $\mathsf{C}^{l,M}_{\beta+1} \in \mathbf{M} = \mathrm{Mod}\left(\mathsf{head}\left(\mathsf{fire}\left(P, \mathrm{Sel}\left(\mathsf{C}^{l,M}, \mathsf{C}^{l,M}_{\beta}\right)\right)\right)\right)$, which shows (a).

Now assume $J \subseteq \mathsf{C}^{l,M}_{\beta+1}$ and $J \in \uparrow \mathsf{C}^{l,M}_{\beta} \cap \mathbf{M}$. Since $J \supseteq \mathsf{C}^{l,M}_{\beta}$ we obtain $J \supseteq \mathsf{C}^{l,M}_{\beta+1}$ by showing $J \setminus \mathsf{C}^{l,M}_{\beta} \supseteq \mathsf{C}^{l,M}_{\beta+1} \setminus \mathsf{C}^{l,M}_{\beta}$. Indeed

$$J \setminus \mathsf{C}^{l,M}_{\beta} \in \left\{K \in \mathbf{I}_P \mid K \models \mathsf{head}\left(R\left(\mathsf{C}^{l,M}_{\beta}, K\right)\right)^{+}\right\}$$

and therefore $J \setminus \mathsf{C}^{l,M}_{\beta} \models \mathsf{head}\left(R\left(\mathsf{C}^{l,M}_{\beta}, J \setminus \mathsf{C}^{l,M}_{\beta}\right)\right)^{+}$. Condition **(L2)**, i.e. minimality of $C_{\beta+1} \setminus \mathsf{C}^{l,M}_{\beta}$ in this set, implies $J \setminus \mathsf{C}^{l,M}_{\beta} \supseteq \mathsf{C}^{l,M}_{\beta+1} \setminus \mathsf{C}^{l,M}_{\beta}$ as desired.

By $J \in \mathrm{Mod}\left(\mathsf{head}\left(\mathsf{fire}\left(P, \mathrm{Sel}\left(\mathsf{C}^{l,M}, \mathsf{C}^{l,M}_{\beta}\right)\right)\right)\right)$ we have $J \models \mathsf{head}\,(r)$ for all rules $r \in \mathsf{fire}\left(P, \mathrm{Sel}\left(\mathsf{C}^{l,M}, \mathsf{C}^{l,M}_{\beta}\right)\right)$. For each of these rules $r$, $J \models \mathsf{head}\,(r)$ is satisfied by $J \models \mathsf{head}\,(r)^{-}$ or by $\mathsf{C}^{l,M}_{\beta} \models \mathsf{head}\,(r)^{+}$ and in both cases we have $r \notin R\left(\mathsf{C}^{l,M}_{\beta}, J\right)$. For all remaining rules, we know that $J \models \mathsf{head}\,(r)$ is satisfied by $J \setminus \mathsf{C}^{l,M}_{\beta} \cap \mathsf{head}\,(r)^{+} \neq \emptyset$, i.e. $J \setminus \mathsf{C}^{l,M}_{\beta} \models \mathsf{head}\,(r)^{+}$, and therefore we know $J \setminus \mathsf{C}^{l,M}_{\beta} \in \left\{K \in \mathbf{I}_P \mid K \models \mathsf{head}\left(R\left(\mathsf{C}^{l,M}_{\beta}, K\right)\right)^{+}\right\}$. By $J \setminus \mathsf{C}^{l,M}_{\beta} \subseteq \mathsf{C}^{l,M}_{\beta+1} \setminus \mathsf{C}^{l,M}_{\beta}$ and minimality of $\mathsf{C}^{l,M}_{\beta+1} \setminus \mathsf{C}^{l,M}_{\beta}$ in the set $\left\{K \in \mathbf{I}_P \mid K \models \mathsf{head}\left(R\left(\mathsf{C}^{l,M}_{\beta}, K\right)\right)^{+}\right\}$ we have $J \setminus \mathsf{C}^{l,M}_{\beta} = \mathsf{C}^{l,M}_{\beta+1} \setminus \mathsf{C}^{l,M}_{\beta}$ and therefore $J = \mathsf{C}^{l,M}_{\beta+1}$, which shows (b).

Thus, $\mathsf{C}^{l,M}_{\beta+1} \in \Omega_P\left(\mathrm{Sel}, \mathsf{C}^{l,M}, \mathsf{C}^{l,M}_{\beta}\right)$. Hence $\mathsf{C}^{l,M}$ is a $(P, M, \mathrm{Sel})$-chain.This proves $M \in \mathrm{Mod}_{\mathrm{Sel}}\,(P)$ and concludes the proof. □

For all selectors Sel, it was shown in [23,24] that the Sel-semantics of programs in GLP is invariant with respect to the following transformations: the replacement ($\rightarrow_{\mathsf{eq}}$) of the body and the head of a rule by logically equivalent formulae and the splitting ($\rightarrow_{\mathsf{hs}}$) of conjunctive heads, more precisely the replacement $P \cup \{\varphi \Rightarrow \psi \wedge \psi'\} \ \rightarrow_{\mathsf{hs}} \ P \cup \{\varphi \Rightarrow \psi, \varphi \Rightarrow \psi'\}$.

Since every formula $\mathsf{head}(r)$ is logically equivalent to a formula in conjunctive normal form, each selector sematics $\mathrm{Mod}_{\mathrm{Sel}}$ of a generalized program $P$ is equivalent to the selector semantics $\mathrm{Mod}_{\mathrm{Sel}}$ of all head disjunctive programs $Q$ where $P \rightarrow^{*}_{\mathsf{eq,hs}} Q$. Note that in the transformation $\rightarrow^{*}_{\mathsf{eq,hs}}$, no shifting of subformulas between the body and the head of a rule is involved. Therefore, Theorem 5 immediately generalizes to the following result.

**Corollary 1.** *Let $P$ be a generalized program and $M$ an interpretation of $P$. Then $M$ is a* Sel*-generated model of $P$ iff for any head disjunctive program $Q$ with $P \rightarrow^{*}_{\mathsf{eq,hs}} Q$ there exists a level mapping $l : \mathsf{B}_Q \rightarrow \alpha$ satisfying **(L1)**, **(L2)** and **(L3)** of Theorem 5.* □

## 6 Corollaries

We can now apply Theorem 5 in order to obtain level mapping characterizations for every semantics generated by a selector, in particular for those semantics generated by the selectors defined in Example 1 and listed in Theorem 1. For syntactically restricted programs, we can furthermore simplify the properties **(L1)**,**(L2)** and **(L3)** in Theorem 5.

### Programs with positive disjunctions in all heads

For rules $r \in \mathsf{HDLP}$, where $\mathsf{head}(r)$ is a disjunction of atoms, we have $\mathsf{head}(r)^- = \emptyset$. Hence we have $\mathsf{head}(r)^- \subseteq I$, i.e. $I \not\models \mathsf{head}(r)^-$, for all interpretations $I \in \mathbf{I}_P$. Thus the set $R\left(\mathsf{C}_\beta^{l,M}, J\right)$ from **(L2)** in Theorem 5 can be specified by

$$R\left(\mathsf{C}_\beta^{l,M}, J\right) = \left\{r \in \mathsf{fire}\left(P, \mathrm{Sel}\left(\mathsf{C}^{l,M}, \mathsf{C}_\beta^{l,M}\right)\right) \mid \mathsf{C}_\beta^{l,M} \not\models \mathsf{head}(r)^+\right\}.$$

We furthermore observe that the set $R\left(\mathsf{C}_\beta^{l,M}, J\right)$ does not depend on the interpretation $J$, so we obtain

$$R'\left(\mathsf{C}_\beta^{l,M}\right) = \left\{r \in \mathsf{fire}\left(P, \mathrm{Sel}\left(\mathsf{C}^{l,M}, \mathsf{C}_\beta^{l,M}\right)\right) \mid \mathsf{C}_\beta^{l,M} \cap \mathsf{head}(r)^+ = \emptyset\right\}$$

and hence

$$\mathrm{Min}\left\{J \in \mathbf{I}_P \,\middle|\, J \models \mathsf{head}\left(R\left(\mathsf{C}_\beta^{l,M}, J\right)\right)^+\right\} = \mathrm{Min}\left(\mathrm{Mod}\left(\mathsf{head}\left(R'\left(\mathsf{C}_\beta^{l,M}\right)\right)\right)\right).$$

Thus for programs containing only rules whose heads are disjunctions of atoms we can rewrite condition **(L2)** in Theorem 5, as follows:

**(L2d)** for every $\beta$ with $\beta + 1 < \alpha$:

$$\mathsf{C}_{\beta+1}^{l,M} \setminus \mathsf{C}_\beta^{l,M} \in \mathrm{Min}\left(\mathrm{Mod}\left(\mathsf{head}\left(R'\left(\mathsf{C}_\beta^{l,M}\right)\right)\right)\right), \text{ where}$$
$$R'\left(\mathsf{C}_\beta^{l,M}\right) = \left\{r \in \mathsf{fire}\left(P, \mathrm{Sel}\left(\mathsf{C}^{l,M}, \mathsf{C}_\beta^{l,M}\right)\right) \,\middle|\, \mathsf{C}_\beta^{l,M} \cap \mathsf{head}(r)^+ = \emptyset\right\}.$$

### Programs with atomic heads

Single atoms are a specific kind of disjunctions of atoms. Hence for programs with atomic heads we can replace condition **(L2)** in Theorem 5 by **(L2d)**, and further simplify it as follows.

For rules with atomic heads we have $\mathsf{head}\left(\{r \in P \mid \mathsf{head}(r) \notin I\}\right) = \mathsf{head}(P) \setminus I$ and therefore

$$\begin{aligned}
&\mathsf{head}\left(R'\left(\mathsf{C}_\beta^{l,M}\right)\right)\\
&= \mathsf{head}\left(\left\{r \in \mathsf{fire}\left(P, \mathrm{Sel}\left(\mathsf{C}^{l,M}, \mathsf{C}_\beta^{l,M}\right)\right) \mid \mathsf{head}(r) \cap \mathsf{C}_\beta^{l,M} = \emptyset\right\}\right)\\
&= \mathsf{head}\left(\left\{r \in \mathsf{fire}\left(P, \mathrm{Sel}\left(\mathsf{C}^{l,M}, \mathsf{C}_\beta^{l,M}\right)\right) \mid \mathsf{head}(r) \notin \mathsf{C}_\beta^{l,M}\right\}\right)\\
&= \mathsf{head}\left(\mathsf{fire}\left(P, \mathrm{Sel}\left(\mathsf{C}^{l,M}, \mathsf{C}_\beta^{l,M}\right)\right)\right) \setminus \mathsf{C}_\beta^{l,M}.
\end{aligned}$$

Because all formulae in $\mathsf{head}(P)$ are atoms we obtain

$$\begin{aligned}\mathrm{Min}\left(\mathrm{Mod}\left(\mathsf{head}\left(R'\left(\mathsf{C}_{\beta}^{l,M}\right)\right)\right)\right) &= \mathrm{Min}\left(\uparrow\left(\mathsf{head}\left(R'\left(\mathsf{C}_{\beta}^{l,M}\right)\right)\right)\right)\\ &= \left\{\mathsf{head}\left(R'\left(\mathsf{C}_{\beta}^{l,M}\right)\right)\right\}\end{aligned}$$

and this allows us to simplify **(L2)** in Theorem 5 to the following:

**(L2a)** for each $\beta$ with $\beta+1<\alpha$:

$$\mathsf{C}_{\beta+1}^{l,M}\setminus\mathsf{C}_{\beta}^{l,M}=\mathsf{head}\left(\mathsf{fire}\left(P,\mathrm{Sel}\left(\mathsf{C}^{l,M},\mathsf{C}_{\beta}^{l,M}\right)\right)\right)\setminus\mathsf{C}_{\beta}^{l,M}.$$

**Inflationary models** From Theorem 1 we know that for normal programs $P$ the selector $\mathrm{Sel}_{\mathsf{l}}$ generates exactly the inflationary model of $P$ as defined in [18]. The generalizations of the definition of inflationary models and this result to head atomic programs are immediate. From [24] we also know that every $\mathrm{Sel}_{\mathsf{l}}$-generated model is generated by a $(P,M,\mathrm{Sel}_{\mathsf{l}})$-chain of length $\omega$. Thus level mappings $l:\mathsf{B}_P\to\omega$ are sufficient to characterize inflationary models of head atomic programs. In this case, condition **(L3)** applies only to the limit ordinal $0<\omega$. But by Remark 1, all level mappings satisfy this property. Therefore we do not need condition **(L3)** in the characterization of inflationary models.

Using Theorem 5 and the considerations above, we obtain the following characterization of inflationary models.

**Corollary 2.** *Let $P\subseteq\mathsf{HALP}$ be a head atomic program and $M$ be an interpretation for $P$. Then $M$ is the inflationary model of $P$ iff there exists a level mapping $l:\mathsf{B}_P\to\omega$ with the following properties.*

**(L1)** $M=\sup\left(\mathsf{C}^{l,M}\right)\in\mathrm{Mod}(P)$.
**(L2i)** *for all $n<\omega$:* $\mathsf{C}_{n+1}^{l,M}\setminus\mathsf{C}_{n}^{l,M}=\mathsf{head}\left(\mathsf{fire}\left(P,\mathsf{C}_{n}^{l,M}\right)\right)\setminus\mathsf{C}_{n}^{l,M}$. □

**Normal programs**

For normal programs, the heads of all rules are single atoms. Hence the simplification **(L2a)** of condition **(L2)** in Theorem 5 applies for all selector generated semantics for normal programs.

The special structure of the bodies of all rules in normal programs allows an alternative formulation of **(L2a)**. In every normal rule, the body is a conjunction of literals. Thus for any set of interpretations $\mathbf{J}$ we have $\mathbf{J}\models\mathsf{body}(r)$ iff $\mathsf{body}(r)^{+}\subseteq J$ and $\mathsf{body}(r)^{-}\cap J=\emptyset$ for all interpretations $J\in\mathbf{J}$.

**Stable models** We develop next a characterization for stable models of normal programs, as introduced in [11]. The selector $\mathrm{Sel}_{\mathsf{lu}}$ generates exactly all stable models for normal programs. In [24], it was also shown that all $\mathrm{Sel}_{\mathsf{lu}}$-generated models $M$ of a program $P$ are generated by a $(P,M,\mathrm{Sel})$-chain of length $\leq\omega$. So for the same reasons as discussed for inflationary models, level mappings with range $\omega$ are sufficient to characterize stable models and condition **(L3)** can be neglected.

For a normal rule $r$ and two interpretations $I, M \in \mathbf{I}_P$ with $I \subseteq M$ we have $\{I, M\} \models \mathsf{body}(r)$, i.e. $I \models \mathsf{body}(r)$ and $M \models \mathsf{body}(r)$, iff $\mathsf{body}(r)^+ \subseteq I$ and $\mathsf{body}(r)^- \cap M = \emptyset$. Combining this with **(L2a)** we obtain the following characterization of stable models for normal programs.

**Corollary 3.** *Let $P \subseteq \mathsf{NLP}$ be a normal program and $M$ an interpretation for $P$. Then $M$ is a stable model of $P$ iff there exists a level mapping $l : \mathsf{B}_P \to \omega$ satisfying the following properties:*

**(L1)** $M = \sup\left(\mathsf{C}^{l,M}\right) \in \mathrm{Mod}(P)$.
**(L2s)** *for all $n < \omega$:*

$$\mathsf{C}^{l,M}_{n+1} \backslash \mathsf{C}^{l,M}_n = \mathsf{head}\left(\left\{r \in P \mid \mathsf{body}(r)^+ \subseteq \mathsf{C}^{l,M}_n, \mathsf{body}(r)^- \cap M = \emptyset\right\}\right) \backslash \mathsf{C}^{l,M}_n. \ \square$$

Comparing this with Theorem 3, we note that both theorems characterize the same set of models. Thus for a model $M$ of $P$ there exists a level mapping $l : \mathsf{B}_P \to \omega$ satisfying **(L1)** and **(L2s)** iff there exists a level mapping $l : \mathsf{B}_P \to \alpha$ satisfying **(Fs)**. The condition imposed on the level mapping in Theorem 3, however, is weaker than the condition in Corollary 3, because level mappings defined by $(P, M, \mathrm{Sel})$-chains are always pointwise minimal.

### Positive head disjunctive programs

For positive head disjunctive programs using $\mathrm{Sel}_{\mathsf{lu}}$, chains of length $\leq \omega$ suffice and condition **(L3)** becomes redundant. We can further refine condition **(L2d)**. Rule bodies are as for normal programs, so so **(L2d)** can be replaced by:

**(L2ds)** for all $n < \omega$:

$$\mathsf{C}^{l,M}_{n+1} \setminus \mathsf{C}^{l,M}_n \in \mathrm{Min}\left(\mathrm{Mod}\left(\mathsf{head}\left(R'\left(\mathsf{C}^{l,M}_n\right)\right)\right)\right), \text{where}$$
$$R'\left(\mathsf{C}^{l,M}_n\right) = \left\{r \mid \mathsf{body}(r)^+ \subseteq \mathsf{C}^{l,M}_n, \mathsf{body}(r)^- \cap M = \emptyset = \mathsf{C}^{l,M}_n \cap \mathsf{head}(r)^+\right\}.$$

We can now easily compare this with Theorem 4 and therefore with the stable model semantics for positive head disjunctive programs. The first and minor difference is as for normal programs, namely that the $\mathrm{Sel}_{\mathsf{lu}}$-characterization forces the considered level mapping to be pointwise minimal. The second difference displays the distinction between the two semantics: While the $\mathrm{Sel}_{\mathsf{lu}}$-semantics enforces minimality at each step in the chain, the stable model semantics does so globally, in a more implicit manner. This causes the connection between stable and $\mathrm{Sel}_{\mathsf{lu}}$-generated models stated in Theorem 1 (4.). We give an example.

*Example 7.* Consider the program $P$ consisting of the following rules.

$$\Rightarrow p \vee q$$
$$p \Rightarrow q$$

Then $P$ has unique stable model $\{q\}$, while both $\{q\}$ and $\{p, q\}$ are $\mathrm{Sel}_{\mathsf{lu}}$-generated models.

**Definite programs**

In order to characterize the least model of definite programs, we can further simplify condition **(L2)** in Theorem 5. Definite programs are a particular kind of head atomic programs. For definite programs, the inflationary and the least model coincide. We can replace condition **(L2)** in Theorem 5 by **(L2i)** in Corollary 2. Since the body of every definite rule is a conjunction of atoms we obtain

$$\mathsf{fire}(P, I) = \left\{ r \in P \mid \mathsf{body}(r)^+ \subseteq I \right\}$$

for every interpretation $I \in \mathbf{I}_P$. Thus we get the following result.

**Corollary 4.** *Let $P \subseteq \mathsf{LP}$ be a definite program and let $M$ be an interpretation for $P$. Then $M$ is the least model of $P$ iff there exists a level mapping $l : \mathsf{B}_P \to \omega$ satisfying the following conditions.*

**(L1)** $M = \sup\left(\mathsf{C}^{l,M}\right) \in \mathrm{Mod}(P)$.
**(L2l)** *for all $n < \omega$:* $\mathsf{C}^{l,M}_{n+1} \setminus \mathsf{C}^{l,M}_{n} = \mathsf{head}\left(\left\{ r \in P \mid \mathsf{body}(r)^+ \subseteq \mathsf{C}^{l,M}_{n} \right\}\right) \setminus \mathsf{C}^{l,M}_{n}$ □

Comparing this to Theorem 2, we note that the relation between the conditions **(L2l)** and **(Fd)** are similar to those of the conditions **(Fs)** und **(L2s)**.

## 7 Related Work

Recently, a number of comparative studies of semantics for logic programs and non-monotonic reasoning have appeared, addressing the lack of systematicity in the subject. We refer to [14,17] for a comprehensive treatment of related work, and only point out some more recent developments.

[19,20] provide studies concerning the definition of semantics which combine the open and the closed world assumption in a flexible way. We believe that their approach can be captured by level mappings in a systematic manner, but details remain to be investigated.

[4,5] use bilattices in the spirit of [10] to arrive at a unified theory encompassing some of the major semantics for non-disjunctive programs. The loosely related [6,7] capture major semantics by means of a framework for inductive definitions. The unification obtained by these approaches is stronger than that by level mappings, but at the loss of flexibility which results in limited applicability to certain semantics only.

[3] provides a comparative study of different modifications of Reiter's default logic, which is related to the stable model semantics. The impact of this work in logic programming, however, remains to be investigated.

## 8 Conclusions and Further Work

Our main result, Corollary 1 respectively Theorem 5 in Section 5, provides a characterization of selector generated models – in general form – by means of level mappings in accordance with the uniform approach proposed in [14,17]. As corollaries from this theorem, we have also achieved level mapping characterizations of several semantics encompassed by the selector generated approach due to [23,24].

Our contribution is technical, and provides a first step towards a comprehensive comparative study of different semantics of logic programs under extended syntax by means of level mapping characterizations. Indeed, a very large number of syntactic extensions for logic programs are currently being investigated in the community, and even for some of the less fancy proposals there is often no agreement on the preferable way of assigning semantics to these constructs.

A particularly interesting case in point is provided by disjunctive and extended disjunctive programs, as studied in [21,12]. While there is more or less general agreement on an appropriate notion of stable model, as given by the notion of *answer set* in [12], there exist various different proposals for a corresponding well-founded semantics, see e.g. [26]. We expect that recasting them by means of level-mappings will provide a clearer picture on the specific ways of modelling knowledge underlying these semantics.

Eventually, we expect that the study of uniform characterizations of different semantics will lead to methods for extracting other, e.g. procedural, semantic properties from the characterizations, like complexity or decidability results.